\documentclass{article}
\usepackage{graphicx}
\usepackage{epstopdf}
\usepackage{lipsum}
\usepackage{float}
\usepackage{spconf,amsmath,epsfig}
\usepackage{multirow}
\usepackage{mathrsfs}
\usepackage{amssymb}
\usepackage[numbers,sort&compress]{natbib}
\usepackage{hyperref} 
\hypersetup{
colorlinks=true,
linkcolor=black,
citecolor=black
}

\usepackage[T1]{fontenc}
\let\OLDthebibliography\thebibliography
\renewcommand\thebibliography[1]{
  \OLDthebibliography{#1}
  \setlength{\parskip}{0pt}
  \setlength{\itemsep}{0pt plus 0.3ex}
}

\begin{document}\sloppy

\def\x{{\mathbf x}}
\def\L{{\cal L}}

\title{MCA: 2D-3D Retrieval with Noisy Labels via Multi-level Adaptive Correction and Alignment}
%
\name{Gui Zou$^1$, Chaofan Gan$^1$, Chern Hong Lim$^2$, Supavadee Aramvith$^3$ and Weiyao Lin$^{1*}$}

\address{$^1$Shanghai Jiao Tong University, China     $^2$Monash University, Malaysia      \\$^3$Chulalongkorn University, Thailand}
\maketitle

\begin{abstract}
With the increasing availability of 2D and 3D data, significant advancements have been made in the field of cross-modal retrieval. Nevertheless, the existence of imperfect annotations presents considerable challenges, demanding robust solutions for 2D-3D cross-modal retrieval in the presence of noisy label conditions. Existing methods generally address the issue of noise by dividing samples independently within each modality, making them susceptible to overfitting on corrupted labels.
To address these issues, we propose a robust 2D-3D \textbf{M}ulti-level cross-modal adaptive \textbf{C}orrection and \textbf{A}lignment framework (MCA). Specifically, we introduce a Multimodal Joint label Correction (MJC) mechanism that leverages multimodal historical self-predictions to jointly model the modality prediction consistency, enabling reliable label refinement. Additionally, we propose a Multi-level Adaptive Alignment (MAA) strategy to effectively enhance cross-modal feature semantics and discrimination across different levels.
Extensive experiments demonstrate the superiority of our method, MCA, which achieves state-of-the-art performance on both conventional and realistic noisy 3D benchmarks, highlighting its generality and effectiveness.

\newcommand\blfootnote[1]{%
\begingroup
\renewcommand\thefootnote{}\footnote{#1}%
\addtocounter{footnote}{-1}%
\endgroup
}
\blfootnote{$^*$Corresponding author. Email: wylin@sjtu.edu.cn}

\end{abstract}

\section{Introduction}
With the rapid growth of 3D data, research on its analysis and applications has gained significant progress. A key application in leveraging 3D data is 2D-3D cross-modal retrieval, which is essential for tasks such as surveillance \cite{li2019gs3d}, virtual reality \cite{fernandez2017access}, and robotics \cite{bellandi2013roboscan}.
Compared to 2D data, 3D data exhibit complex geometric structures, high-dimensional representations, and irregular distributions, making their processing more challenging. Moreover, the rapid expansion of 3D datasets has made annotation increasingly costly and labor-intensive, leading to prevalent label noise in cross-modal retrieval tasks. Consequently, learning effectively under noisy labeling conditions has become a critical challenge in advancing 2D-3D cross-modal retrieval techniques.

Previous studies \cite{feng2023rono,hu2021learning} on cross-modal retrieval with noisy labels generally follow two steps: mitigating the adverse effects of noisy labels and bridging the heterogeneity gap between multimodal data.
To address label noise, various methods such as GCE \cite{zhang2018generalized}, Label Smoothing \cite{lukasik2020does}, and DivideMix \cite{li2020dividemix} have been proposed, demonstrating effectiveness in unimodal scenarios. Meanwhile, to reduce the heterogeneity gap in multimodal data, approaches leveraging center loss \cite{jing_cross-modal_2021} have been explored, aiming to align cross-modal features while minimizing intra-class variations. These methods have proven effective in learning a shared feature space when well-annotated data are available.

However, conventional noise-robust learning methods \cite{ghosh2017robust, han2019deep} are primarily designed for unimodal tasks and struggle to extend to 2D-3D multimodal domains due to the substantial heterogeneity and domain gap between 2D and 3D data. Meanwhile, traditional supervised cross-modal alignment approaches \cite{zhen2019deep} are highly sensitive to label noise and prone to overfitting on corrupted annotations. Recently, DAC \cite{gan2024dac} introduced a multimodal dynamic division strategy that partitions noisy datasets based on the multimodal sample loss distribution. However, this method overlooks the interaction between different modality outputs, making it susceptible to the choice of division thresholds.

To address these challenges, we propose a robust 2D-3D Multi-level Cross-modal Adaptive Correction and Alignment framework (MCA), which consists of Multimodal Joint label Correction (MJC) and Multi-level Adaptive Alignment (MAA). Specifically, MJC utilizes multimodal historical self-predictions to reliably refine noisy labels, allowing the model to extract valuable semantic information from the noisy data. With the corrected labels, different samples are dynamically processed using varying alignment strategies: center-level, group-level, and instance-level alignments. Through this adaptive alignment approach, MCA effectively captures both semantic and intrinsic geometric information across modalities, accommodating varying noise ratios and facilitating the learning of robust, discriminative, and modality-invariant representations.
Our key contributions are summarized as follows:

\begin{itemize}
\item 
We propose a novel and robust 2D-3D Multi-level cross-modal adaptive Correction and Alignment framework (MCA) that effectively corrects noisy labels and adaptively align the cross-modal features. 

\item 
In MCA, we introduce Multimodal Joint Label Correction (MJC), which leverages multimodal information to reliably refine noisy labels and enhance feature quality.

\item 
In MCA, we propose a Multi-level Adaptive Alignment (MAA) strategy to effectively align cross-modal features across multiple levels while mitigating the adverse effects of label noise.

\item 
Our method exhibits remarkable effectiveness and robustness, achieving state-of-the-art performance across both traditional and real-world 3D object benchmarks.

\end{itemize}

\section{Related Work}
\subsection{Cross-modal Retrieval}
Cross-modal retrieval methods aim to unify heterogeneous modalities within a shared representation space. Some approaches achieve this by leveraging statistical dependencies, such as Canonical Correlation Analysis (CCA) \cite{wang2015deep} and Multi-View Learning \cite{kan2015multi}. More recently, methods like contrastive learning \cite{hu2021learning} enhance alignment by maximizing mutual information between paired data.
Other strategies rely on explicit supervision to enforce modality-invariant representations. For example, \cite{zhen2019deep} optimizes both label and feature space alignment, while metric learning techniques \cite{jing_cross-modal_2021} use center loss to refine intra- and inter-class relationships. To handle noisy annotations, \cite{gan2024dac} introduces a robust loss function. However, most existing methods assume clean data, limiting their effectiveness in 2D-3D retrieval with noisy labels.

\subsection{Learning with Noisy Labels}
Learning with noisy labels is a critical challenge in machine learning, with existing methods typically falling into three categories: robust loss functions, label correction, and sample selection.
Robust loss functions mitigate the impact of label noise by designing noise-tolerant objectives, reducing overfitting on corrupted labels. Label correction techniques refine noisy annotations through model-driven self-correction or external meta-correctors for higher-quality adjustments \cite{zheng2021meta}. Sample selection approaches, on the other hand, aim to identify clean instances within noisy datasets. The widely used small-loss trick selects low-loss samples as reliable data points \cite{han2018co, shen2019learning}, while Dividemix \cite{li2020dividemix} leverages a Gaussian Mixture Model (GMM) to further distinguish clean and noisy instances. Additionally, historical model predictions offer an alternative selection criterion, enriching the decision process and improving reliability \cite{wei2022self, feng2023rono}. Recently, DAC~\cite{gan2024dac} introduced a multimodal division strategy that dynamically separates clean and noisy samples by modeling the multimodal loss distribution. However, it overlooks the intrinsic relationships between different modalities. This limitation motivates us to develop a more reliable approach that not only distinguishes samples effectively but also refines noisy labels by deeply modeling cross-modal relationships.

\begin{figure*}[tb]
  \centering
  \includegraphics[width = \linewidth]{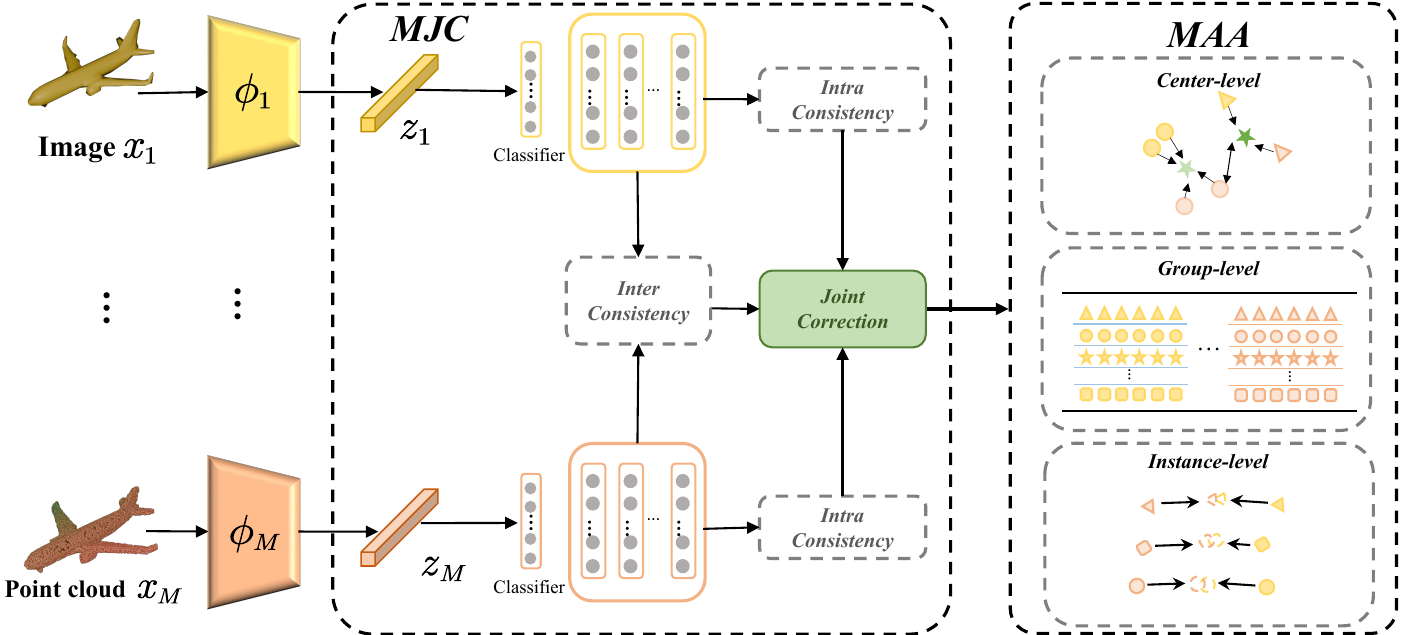}
  \caption{An overview of our proposed MCA framework. Our model first refines noisy labels using multimodal historical self-predictions through our MJC mechanism. It then adaptively enhances cross-modal feature semantics and discrimination through our MAA strategy.
  }
  \label{figure:framework}
  \vspace{-4mm}
\end{figure*}
\section{Methodology}

\subsection{Overview}
Following the setting in \cite{gan2024dac}, we define the 2D-3D multimodal dataset as:
$
\mathcal{D} = \left\{\mathcal{M}_j\right\}_{j=1}^M = \left\{\mathcal{X}_j, \mathcal{Y}_j\right\}_{j=1}^M
$
where \( M \) represents the number of modalities, and the dataset consists of \( K \) distinct categories.  
For each modality \( \mathcal{M}_j \), the dataset contains \( N \) samples, formulated as:
$
\mathcal{M}_j = \left\{\left(\boldsymbol{x}_i^j, y_i^j\right)\right\}_{i=1}^N
$
where \( \boldsymbol{x}_i^j \) denotes the \( i \)-th sample in the \( j \)-th modality, and \( y_i^j \) is its corresponding noisy label. For each sample $x_i^j$, we extract the sample feature $z_i^j=\phi_j(x_i^j)$ with the corresponding feature extractor $\phi_j$. For simplicity, we denote $(x_j,y_j)=(x_i^j,y_i^j)$ in the following sections. The overview of our framework MCA is shown in Fig.~\ref{figure:framework}.


\subsection{MJC: Multimodal Joint label Correction}
\label{sec:MJC}
To address label noise, we propose a Multimodal Joint Label Correction (MJC) strategy to reliably refine corrupted labels and prevent models from overfitting to erroneous annotations. Specifically, MJC performs joint label correction by leveraging multimodal self-predictions \cite{wei2022self, bai2021me}. As illustrated in Fig.~\ref{figure:framework}, we utilize self-predictions from different modalities to jointly estimate sample credibility and correct corrupted labels.
However, relying solely on self-predictions from a single epoch is unreliable due to training instability. To enhance robustness, we design MJC using multiple historical predictions, as depicted in Fig.~\ref{figure:framework}. We maintain a historical bank $\mathcal{H}_i^j$
that stores each sample’s historical single-modality self-predictions. By analyzing prediction temporal consistency, we compute intra-modal consistency to generate more reliable corrected labels.
Additionally, to mitigate the confirmation bias inherent in single-modality predictions, we introduce inter-modal consistency between different modalities. This cross-modal constraint effectively reduces modality-specific biases, leading to more accurate and robust label correction.

\noindent\textbf{Historical bank.}
Specifically, the self-prediction of the model for each sample is denoted as $p_{i,m}^j$, where $m$ denotes the epoch. The historical bank of $j$-th modality is denoted as $\mathcal{H_i^j}$ and the size of the historical bank is $h$. For the sample $(x_i^j, p_{i,m}^j)$, we update the $\mathcal{H}_i^j$ by its self-prediction $p_{i,m}^j$ via the First-in First-out principle. The historical bank for each modality is formulated as follows:
\begin{equation}
\mathcal{H}_i^j = [p_{i,(m-h+1)}^j, p_{i,(m-h+2)}^j, ... , p_{i,m}^j]_i
\end{equation}
where $p_{i,m}^j$ is the probability prediction of sample in epoch $m$, and $i$ is the index number of historical bank.

\noindent\textbf{Intra and inter Consistency.}
Next, we leverage the temporal self-predictions over $h$ epochs to compute both intra-modal and inter-modal consistency for each sample. Specifically, intra-modal consistency measures the frequency of each label appearing in the historical bank $\mathcal{H}_i^j$. We select the label with the highest occurrence as the predicted label for modality $j$.
\begin{equation}
C_i^j =unique(\mathcal{H}_i^j)
\end{equation}
where $C_i^j$ represents the predicted label derived from the historical bank $\mathcal{H}_i^j$ and $unique$ denotes a counting function that determines the most frequently occurring label. 

Furthermore, inter-modal consistency is computed by jointly considering the predicted labels across multiple modalities, formulated as $\cap\{C_i^j\}$, where the intersection operation ensures that the final label prediction is consistent across modalities, enhancing robustness against noise.

\noindent\textbf{Joint Correction.}
With the intra and inter consistency, the joint correction criterion $H$ of each sample is formulated as $\cap\{C_i^j\}$,
\begin{equation}
H(x_i^j) = \cap\{C_i^j\}
\end{equation}
Finally, we utilize the original labels $y_i^j$ to divide the samples into clean set $S_c$ and noisy set $S_n$ :
\begin{equation}
S_c=\{x_i|H(x_i^j)=y_i^j\}, S_n=\{x_i|H(x_i^j)\neq y_i^j\}
\end{equation}

\subsection{MAA: Multi-level Adaptive Alignment}
\label{sec:MAA}
As shown in Fig. \ref{figure:framework}, to fully harness the discriminative information in noisy samples and meanwhile mitigate the negative effects from noisy labels, we design a Multi-level cross-modal Adaptive Alignment (MAA) to dynamically align cross-modal features from multiple views to capture different degrees of semantic and geometric information. Specifically, we dynamically align the cross-modal from three views (center-level, group-level, and instance-level) to eliminate the semantic gap across multiple modalities.


\begin{table*}[htb]
\setlength\tabcolsep{3pt}
\centering
\caption{Performance comparison for image-to-point cloud (Img → Pnt) and point cloud-to-image (Pnt → Img) retrieval on ModelNet10 and ModelNet40 under symmetric noise rates of 0.2, 0.4, 0.6, and 0.8. The best and second-best results are highlighted in bold and underlined, respectively.}
\label{tab:symmetric}
\resizebox{\linewidth}{!}{
\begin{tabular}{|c|cccc|cccc|cccc|cccc|}
\hline & \multicolumn{8}{c|}{ ModelNet10 \cite{wu20153d} } & \multicolumn{8}{c|}{ ModelNet40 \cite{wu20153d} } \\
\cline { 2 - 17 } Method & \multicolumn{4}{c|}{ Img $\rightarrow$ Pnt } & \multicolumn{4}{c|}{ Pnt $\rightarrow$ Img } & \multicolumn{4}{c|}{ Img $\rightarrow$ Pnt } & \multicolumn{4}{c|}{ Pnt $\rightarrow$ Img } \\
& 0.2 & 0.4 & 0.6 & 0.8 & 0.2 & 0.4 & 0.6 & 0.8 & 0.2 & 0.4 & 0.6 & 0.8 & 0.2 & 0.4 & 0.6 & 0.8 \\
\hline CCA \cite{hotelling1992relations} & 0.625 & 0.625 & 0.625 & 0.625 & 0.627 & 0.627 & 0.627 & 0.627 & 0.532 & 0.532 & 0.532 & 0.532 & 0.531 & 0.531 & 0.531 & 0.531 \\
DGCPN \cite{yu2021deep} & 0.765 & 0.765 & 0.765 & 0.765 & 0.759 & 0.759 & 0.759 & 0.759 & 0.705 & 0.705 & 0.705 & 0.705 & 0.697 & 0.697 & 0.697 & 0.697 \\
UCCH \cite{hu2022unsupervised} & 0.771 & 0.771 & 0.771 & 0.771 & 0.770 & 0.770 & 0.770 & 0.770 & 0.755 & 0.755 & 0.755 & 0.755 & 0.739 & 0.739 & 0.739 & 0.739 \\
\hline
\hline
DAGNN \cite{qian2021dual} & 0.844 & 0.800 & 0.754 & 0.422 & 0.836 & 0.810 & 0.763 & 0.448 & 0.802 & 0.723 & 0.635 & 0.402 & 0.798 & 0.728 & 0.643 & 0.412 \\
ALGCN \cite{qian2021adaptive} & 0.788 & 0.597 & 0.426 & 0.282 & 0.797 & 0.589 & 0.440 & 0.269 & 0.766 & 0.538 & 0.426 & 0.298 & 0.763 & 0.537 & 0.403 & 0.279 \\
DSCMR \cite{zhen2019deep} & 0.849 & 0.758 & 0.666 & 0.324 & 0.836 & 0.732 & 0.637 & 0.307 & 0.824 & 0.788 & 0.687 & 0.328 & 0.811 & 0.785 & 0.694 & 0.339 \\
MRL \cite{hu2021learning} & 0.876 & 0.870 & 0.863 & 0.832 & 0.861 & 0.857 & 0.848 & 0.823 & 0.833 & 0.829 & 0.828 & 0.818 & 0.824 & 0.826 & 0.820& 0.817 \\
CLF \cite{jing_cross-modal_2021} & 0.849 & 0.782 & 0.620 & 0.365 & 0.838 & 0.764 & 0.595 & 0.387 & 0.822 & 0.778 & 0.624 & 0.315 & 0.815 & 0.771 & 0.587 & 0.295 \\
CLF+MAE\cite{ghosh2017robust} & 0.853 & 0.752 & 0.679 & 0.343 & 0.838 & 0.716 & 0.659 & 0.373 & 0.827 & 0.758 & 0.651 & 0.384 & 0.816 & 0.749 & 0.640 & 0.372 \\
RONO \cite{feng2023rono} & 0.892 & 0.877 & 0.870 & 0.836 & 0.890 & 0.875 & 0.861 & 0.830 & 0.877 & 0.858 & 0.838 & 0.823 & 0.872 & 0.854 & 0.838 & 0.821 \\
DAC~\cite{gan2024dac} & $\mathbf{0.898}$ & $\underline{0.897}$ & $\underline{0.890}$ & $\underline{0.879}$ & $\underline{0.901}$ & $\underline{0.895}$ & $\underline{0.888}$ & $\underline{0.881}$ & $\underline{0.894}$ & $\underline{0.893}$ & $\underline{0.893}$ & $\underline{0.879}$ & $\mathbf{0.886}$ & $\underline{0.885}$ & $\underline{0.884}$ & $\underline{0.871}$ \\
\hline
MCA(Ours) & $\underline{0.893}$ & $\mathbf{0.902}$ & $\mathbf{0.895}$ & $\mathbf{0.880}$ & $\mathbf{0.901}$ & $\mathbf{0.902}$ & $\mathbf{0.889}$ & $\mathbf{0.887}$ & $\mathbf{0.897}$ & $\mathbf{0.897}$ & $\mathbf{0.895}$ & $\mathbf{0.886}$ & $\underline{0.882}$ & $\mathbf{0.895}$ & $\mathbf{0.889}$ & $\mathbf{0.878}$ \\
\hline 
\end{tabular}}
\vspace{-4mm}
\end{table*}

\noindent\textbf{Center-level alignment.} It aims to align cross-modal representations within a globally shared space, effectively minimizing intra-class variation across multiple modalities. Specifically, it pulls samples with the same categories toward the corresponding global clustering center while pushing them away from other clustering centers, which eliminates intra-class semantic discrepancies across multimodal features from a holistic perspective. We adopt a multimodal center loss to achieve the center-level alignment, which is formulated as:
\begin{equation}
L_{c}=-\frac{1}{M N}\sum_i^N \sum_j^M \log\left(\frac{e^{\frac{1}{{\tau}_c}\left(\boldsymbol{c}_k\right)^T \boldsymbol{z}_i^j}}{\sum_{n=1}^Ke^{\frac{1}{{\tau}_c}\left(\boldsymbol{c}_n\right)^T \boldsymbol{z}_i^j}}\right)
\label{loss:center}
\end{equation}
where $c_k$ is the shared clustering center of categories $y_k$ in the common space, ${\tau}_c$ is a temperature parameter.

\noindent\textbf{Group-level alignment.}
In order to capture the individual semantic and geometric structure information of each sample, we introduce group-level alignment, which aims to align the cross-modal feature at a group level. Specifically, we first construct a multimodal semantic group $\mathcal{G} \{G_i\}_{i=1}^K$ where $G_i$ is the group of $i$-th category, and the size of $G_i$ is $M*D$, where $M$ is the number of modalities and $D$ is the group size of each modality to store the cross-modal feature guided by the categories of each sample. For the sample $(x_i^j, y_i^j)$, we update the $\mathcal{G}_i$ by its feature embedding $z_i^j$ via the First-in First-out principle. After finishing the construction of the multimodal semantic group, we can apply the group-level alignment to the noisy data. The cross-modal group-level alignment loss is formulated as:
\begin{equation}
\mathcal{L}_{g}=-\frac{1}{M N} \sum_i^N \sum_j^M \log \left(\frac{\sum_{z_i^k\in G_{y_i}} e^{\frac{1}{{\tau}_m}\left(\boldsymbol{z}_i^k\right)^T \boldsymbol{z}_i^j}}{ \sum_{G_l\in \mathcal{G}}\sum_{z_l^m\in G_l} e^{\frac{1}{{\tau}_m}\left(\boldsymbol{z}_l^m\right)^T \boldsymbol{z}_i^j}}\right)
\end{equation}
where $G_{y_i}$ is the group of category $y_i$.

\noindent\textbf{Instance-level alignment.}
Specifically, we adopt a Multi-modal Modal Gap loss(MG) \cite{hu2021learning} to optimize the cross-modal representation of samples in instance level. The Instance-level alignment loss is formulated as:
\begin{equation}
\mathcal{L}_{i}=-\frac{1}{M N} \sum_i^N \sum_j^M \log \left(\frac{\sum_k^M e^{\frac{1}{{\tau}_m}\left(\boldsymbol{z}_i^k\right)^T \boldsymbol{z}_i^j}}{\sum_l^N \sum_m^M e^{\frac{1}{{\tau}_m}\left(\boldsymbol{z}_l^m\right)^T \boldsymbol{z}_i^j}}\right)
\end{equation}

For the clean set $S_c$, we apply all three levels of alignment strategies to enhance feature quality effectively.
For the noisy set $S_n$, where label corruption is significant, we adopt only group-level and instance-level alignment to mitigate the inherent modality gap, where we utilize the corrected label $H(x_i^j)$ for training.

Additionally, to alleviate the semantic discrepancy of cross-modal representation in the label space, we also adopt the classifier loss to optimize the feature in the label space. For clean set $S_c$, we utilize the conventional cross-entropy (CE) loss $L_c$, which is formulated as:
\begin{equation}
\mathcal{L}_{cls}=-\frac{1}{M N} \sum_i^N \sum_j^My_i\log(y_i^j)
\end{equation}
\noindent\textbf{Overall loss.} The complete loss function of our MCA is formulated as:

\begin{equation}
\mathcal{L}= (\mathcal{L}_c+\mathcal{L}_g+\mathcal{L}_i)+\lambda\mathcal{L}_{cls}
\end{equation}
where $\lambda$ represents the balance parameters for the losses.

\section{EXPERIMENTALS}
\subsection{Experimental Setup}
\noindent\textbf{Noisy test benchmarks.}
We conduct extensive experiments on both traditional 3D datasets ModelNet10 and ModelNet40~\cite{wu20153d} and the realistic noisy 3D dataset Objaverse-N200~\cite{gan2024dac}.
For ModelNet10/40, we follow the noisy benchmark construction proposed in~\cite{gan2024dac}, considering the symmetric noise setting with noise rates of 0.2, 0.4, 0.6, and 0.8. Additionally, Objaverse-N200~\cite{gan2024dac} serves as a newly introduced realistic noisy 3D benchmark, provided by DAC~\cite{gan2024dac}.
We perform comprehensive evaluations on these datasets to assess the effectiveness and robustness of our proposed model.

\noindent\textbf{Implementation details.}
For feature extraction, we use ResNet18~\cite{he2016deep} for RGB images, DGCNN~\cite{wang2019dynamic} for 3D point clouds, and MeshNet~\cite{feng2019meshnet} for 3D meshes on traditional datasets, projecting features into a 512D space with two fully connected layers. For Objaverse-N200, we adopt EVA-CLIP~\cite{sun2023eva} for RGB images and Uni3D~\cite{zhou2023uni3d} for point clouds, mapping features to 1024D. 
The training is done using Adam optimizer [32] (momentum 0.9, batch size 64), and it is trained for 400 epochs (lr=0.0001, decayed every 50 epochs). For objaverse-N200, it is trained for 40 epochs (lr=0.0001, decayed at epochs 20,40).

\subsection{Results}
\noindent\textbf{Results on traditional benchmarks.}
We first evaluate our proposed MCA framework on the traditional noisy 3D datasets ModelNet10 and ModelNet40~\cite{wu20153d}, with the results summarized in Table~\ref{tab:symmetric}. As observed, MCA consistently outperforms previous state-of-the-art methods across different noise levels, demonstrating its effectiveness in robust 2D-3D cross-modal retrieval. Specifically, under 40\% symmetric noise on ModelNet40, MCA achieves a retrieval performance of 89.7\%, marking a significant improvement of 0.4\% $\uparrow$ over existing approaches. Furthermore, our model exhibits strong resilience to label noise, maintaining high retrieval accuracy across a wide range of noise rates. Even when the noise ratio increases from 20\% to 60\%, MCA achieves stable and competitive results (ranging from 89.5\% to 89.7\%), highlighting its robustness in handling noisy multimodal datasets.

\begin{table}
\setlength\tabcolsep{3pt}
\centering
\caption{Performance comparison on the real-world dataset Objacerse-N200.}
\label{tab:objaverse}
\resizebox{0.75\linewidth}{!}{
\begin{tabular}{|c|c|c|}
\hline 
\multirow{2}{*}{Method} & \multicolumn{2}{c|}{ Objaverse-N200}\\
\cline { 2 - 3 } & \multicolumn{1}{c|}{ Img $\rightarrow$ Pnt } & \multicolumn{1}{c|}{ Pnt $\rightarrow$ Img }\\
\hline
Openshape\cite{liu2024openshape}& 0.274 & 0.257\\
\hline
MRL\cite{hu2021learning} &0.260 &0.252\\
CLF\cite{jing_cross-modal_2021} & 0.238 & 0.192\\
CLF\cite{jing_cross-modal_2021}+MAE\cite{ghosh2017robust} &0.237 &0.192 \\
RONO\cite{feng2023rono} & 0.276 & 0.279 \\
DAC~\cite{gan2024dac} & 0.334 & 0.338 \\
\hline
MCA & $\mathbf{0.343}$ & $\mathbf{0.341}$ \\
\hline

\end{tabular}}
\vspace{-4mm}
\end{table}

\begin{table}[tb]
\setlength\tabcolsep{3pt}
\caption{Ablation studies for MJC on the on the ModelNet10/40 datasets with 0.4 symmetric noise. }
\label{tab:MJC}
\centering
\resizebox{\linewidth}{!}{
\begin{tabular}{|cc|c|c|c|c|}
\hline \multicolumn{2}{|c|}{MJC} & \multicolumn{2}{c|}{ ModelNet10 \cite{wu20153d} } & \multicolumn{2}{c|}{ ModelNet40 \cite{wu20153d} } \\
\hline intra&inter &Img $\rightarrow$ Pnt& Pnt $\rightarrow$ Img & Img $\rightarrow$ Pnt & Pnt $\rightarrow$ Img \\
\hline $\checkmark$&$\checkmark$ & $\mathbf{0.902}$&$\mathbf{0.902}$ & $\mathbf{0 . 8 97}$& $\mathbf{0 . 8 9 5}$ \\
$\checkmark$& & 0.893&0.891  & 0.871 &0.882 \\
& $\checkmark$ & 0.891 &0.887 & 0.870&0.879 \\
\hline
\end{tabular}}
\vspace{-4mm}
\end{table}

\noindent\textbf{Results on real-world benchmarks.}
To further evaluate the effectiveness and robustness of our MCA framework in real-world scenarios, we conduct extensive experiments on the Objaverse-N200 dataset~\cite{gan2024dac}, which contains realistic noisy 3D data. As shown in Table~\ref{tab:objaverse}, MCA consistently achieves superior retrieval performance(34.3\% vs 33.4\%), demonstrating its capability to handle complex noise patterns more effectively than existing methods. Notably, our model not only outperforms previous approaches in terms of retrieval accuracy but also exhibits strong resilience to real-world label noise, maintaining stable and high-quality feature representations across different levels of corruption. These results further validate the adaptability and robustness of MCA in practical multimodal retrieval tasks.

\noindent\textbf{Ablation Study.}
To evaluate the effectiveness of our proposed strategies, MJC and MAA, we conduct comprehensive ablation studies on the ModelNet10/40 datasets under 40\% symmetric noise. From the results in Table.~\ref{tab:MJC}, it can be observed that our Multimodal Joint label Correction (MJC) achieves a 2.6\% improvement by leveraging inter-modal consistency to model multimodal feature interactions, highlighting the crucial role of complementary information across modalities in label correction. Additionally, the results in Table.~\ref{tab:MAA} reveal that our Multi-level Adaptive Alignment (MAA) effectively addresses noise in 2D-3D retrieval. Each component of MAA contributes significantly to bridging the cross-modal feature gap, enhancing feature semantics, and improving discriminability. These findings confirm the reliability and effectiveness of our MJC and MAA strategies.

\begin{table}[tb]
\setlength\tabcolsep{3pt}
\caption{Ablation studies for MAA on the on the ModelNet10/40 datasets with 0.4 symmetric noise.}
\label{tab:MAA}
\centering
\resizebox{\linewidth}{!}{
\begin{tabular}{|ccc|c|c|c|c|}
\hline \multicolumn{3}{|c|}{MAA} & \multicolumn{2}{c|}{ ModelNet10 \cite{wu20153d} } & \multicolumn{2}{c|}{ ModelNet40 \cite{wu20153d} } \\
\hline $S_c$&$S_g$&$S_i$ &Img $\rightarrow$ Pnt& Pnt $\rightarrow$ Img & Img $\rightarrow$ Pnt & Pnt $\rightarrow$ Img \\
\hline $\checkmark$&$\checkmark$&$\checkmark$ & $\mathbf{0.902}$&$\mathbf{0.902}$ & $\mathbf{0 . 8 97}$& $\mathbf{0 . 8 9 5}$ \\
&$\checkmark$&$\checkmark$& 0.897&0.896  & 0.887 &0.882 \\
$\checkmark$&&& 0.823 &0.815 & 0.801&0.815 \\
&$\checkmark$& & 0.836 &0.827 & 0.823&0.837 \\
&&$\checkmark$ & 0.876 &0.875 & 0.870&0.876 \\
\hline
\end{tabular}}
\vspace{-4mm}
\end{table}

\section{Conclusion}
In this paper, we introduced a novel and robust 2D-3D Multi-level cross-modal Correction and Alignment framework (MCA) to address the challenges of cross-modal learning with noisy labels. Specifically, we propose a Multimodal Joint Label Correction strategy to effectively refine mislabeled data, mitigating the adverse impact of label noise while extracting meaningful semantics from the corrupted data. Furthermore, a Multi-level Adaptive Alignment strategy is devised to compactly align different modalities within a shared semantic space, enhancing feature semantics and discriminability. Extensive experiments on both synthetic and real-world noisy 3D datasets validate the effectiveness and robustness of our approach.
\section{Acknowledgement}
The paper is supported in part by the National Natural Science
Foundation of China (No. 62325109, U21B2013), and in part by the  Shanghai ’The Belt and Road’ Young Scholar Exchange Grant (24510742000).

\bibliographystyle{IEEEbib}
\bibliography{icme2022template_short}

\end{document}